\documentclass[sigchi]{acmart}
\usepackage[utf8x]{inputenc}
\usepackage{wrapfig}
\usepackage{scalefnt}
\def\BibTeX{{\rm B\kern-.05em{\sc i\kern-.025em b}\kern-.08emT\kern-.1667em\lower.7ex\hbox{E}\kern-.125emX}}
    
\setcopyright{acmcopyright}
\begin{document}
\copyrightyear{2020}
\acmYear{2020}
\acmConference[ICONS '20]{International Conference on Neuromorphic Systems}{July 28--30, 2020}{Virtual Conference}
\acmBooktitle{International Conference on Neuromorphic Systems (ICONS '20), July 28--30, 2020, Virtual Conference}
\acmPrice{--.00}
\acmDOI{--.--/------.----}
\acmISBN{------/--/--}
%
\title[SIT Learning with Features Extracted from an SCNN]{Continuous Learning in a Single-Incremental-Task Scenario with Spike Features }

%
\author{Ruthvik Vaila}
\email{ruthvikvaila@boisestate.edu}
\author{John Chiasson}
\email{johnchiasson@boisestate.edu}
\affiliation{%
  \institution{Boise State University}
  \streetaddress{ 1910 W University Dr}
  \city{Boise}
  \state{Idaho}
  \country{USA}
  \postcode{83706}
}

\author{Vishal Saxena}
\email{vsaxena@udel.edu}
\affiliation{%
  \institution{University of Delaware}
  \streetaddress{140 Evans Hall}
  \city{Newark}
  \state{DE}
  \country{USA}
  \postcode{19716}
}

%
\renewcommand{\shortauthors}{Vaila, Chiasson, Saxena.}

%
\begin{abstract}
Deep Neural Networks (DNNs) have two key deficiencies, their dependence on high precision computing and their inability to perform sequential learning, that is, when a DNN is trained on a first task and the same DNN is trained on the next task it forgets the first task. This phenomenon of forgetting previous tasks is also referred to as catastrophic forgetting. On the other hand a mammalian brain outperforms DNNs in terms of energy efficiency and the ability to learn sequentially without catastrophically forgetting. Here, we use bio-inspired Spike Timing Dependent Plasticity (STDP) in the feature extraction layers of the network with instantaneous neurons to extract meaningful features. In the classification sections of the network we use a modified synaptic intelligence that we refer to as cost per synapse metric as a regularizer to immunize the network against catastrophic forgetting in a Single-Incremental-Task scenario (SIT). In this study, we use MNIST handwritten digits dataset that was divided into five sub-tasks.
\end{abstract}

%
%


%

\maketitle
\vspace{-0.4cm}
\section{Introduction}
Typically, Spiking Neural Networks (SNNs) are trained using an unsupervised algorithm called Spike Timing Dependant Plasticity (STDP) \cite{Kheradpisheh_2016}. Spike features extracted from latency encoded convolutional variants of SNNs have been used with an SVM \cite{Kheradpisheh_2016} and a linear neural network classifier \cite{vaila2019a} to achieve classification accuracies in excess of $98.5\%$. However, SNNs tend to achieve lower classification accuracies when compared to Artificial Neural Networks (ANNs) \cite{panda2019}. ANNs are trained using Stochastic Gradient Descent (SGD). The main assumption of SGD is that the mini-batches of the training data contain approximately equal number of data points with the same labels (i.e., the data is uniformly randomly distributed). This assumption does not hold for many of the machine learning systems that learn online continuously. Different kinds of continuous learning schemes have been proposed to mitigate the problem of catastrophic forgetting. Two main scenarios of continuous learning are the Multi-Task (MT) and the Single-Incremental-Task (SIT) scenarios \cite{MALTONI2019}. In the MT scenario a neural network with a disjoint set of output neurons is used to train/test a corresponding set of disjoint tasks. In contrast, a neural network for the SIT scenario expands the number of neurons in the output layer to accommodate new classification tasks. The MT scenario is useful when training different classification tasks on the same network thereby allowing resource sharing. The SIT scenario is useful for online continuous learning applications. That is, the SIT scenario is more suitable for online machine learning systems and is more difficult compared to the MT scenario. This is because the MT network has to not only mitigate catastrophic forgetting, but also learn to differentiate classes that are usually not seen together (unless the system has some kind of short term memory to be replayed later). Self-Organizing Maps (SOM) with short-term memory were used in \cite{Gepperth2016} \cite{parisi2019} to achieve an accuracy of $85\%$ on the MNIST dataset using a SIT scenario and replaying the complete dataset. Using STDP based unsupervised learning and dopaminergic plasticity modulation controlled forgetting was proposed in \cite{allred2019}. It was shown to achieve a $~95\%$ accuracy on MNIST dataset using the SIT scenario. Unsupervised spiking networks with predictive coding have been trained with STDP and shown to achieve an accuracy of $~76\%$ on the MNIST dataset using the MT scenario \cite{ororbia2019}. In our work here, our network classifies the data according to the AR1 method given in \cite{MALTONI2019}. This uses the SIT scenario which was inspired by synaptic intelligence for the MT scenario in \cite{zenke2017}. In our previous work \cite{vaila2019a} we used the MNIST dataset split into two disjoint tasks to show that features extracted from a spiking convolutional network (SCN) demonstrated more immunity to catastrophic forgetting compared to their ANN counterparts. In \cite{vaila2019a}, using early stopping, the first five output neurons were trained to classify the digits $\{0,1,2,3,4\}$ and then the remaining five output neurons were trained to classify the digits $\{5,6,7,8,9\}$. The network was then tested on the complete test dataset (digits 0-9) and achieved a $93\%$ accuracy on this test data. In the work presented here we exclusively work with spike features extracted from an SCN and study the effect of continuous learning using the SIT scenario on the MNIST dataset. For this study the MNIST dataset was split into the five disjoint classification tasks $\{\{0,1\},\{2,3\},\{4,5\},\{6,7\},\{8,9\}\}$. The feature classification is done unsupervised in the convolution layer (L2) while the classification is done in the latter layers using error backpropagation. Here we modify the synaptic intelligence regularizer calculation of \cite{zenke2017} in order to reduce the computational load.  
\vspace{-0.6cm}
\section{Network}
\begin{center}
\includegraphics[width=0.8\linewidth, height=0.4\linewidth]{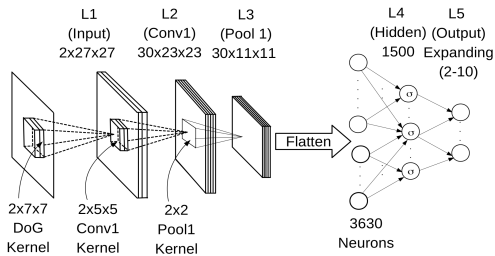}
\captionof{figure}{Layers L1-L3 and L3-L5 are feature extraction and feature classification layers respectively. The output layers expands from $2-10$ output neurons to accommodate the five classification tasks.}
\label{network1}
\end{center}

The feature extraction part of the network is same as in \cite{vaila2019} \cite{vaila2019a}. Input images are encoded into spikes using ON and OFF center DoG filters followed by thresholding \cite{Kheradpisheh_2016}. The L2 (convolution) layer consists of 30 maps and the neurons that emerge as winners after lateral inhibition and STDP competition \cite{vaila2020} get to update their weights according to a simplified STDP \cite{Kheradpisheh_2016}.   
\begin{equation}
\scalefont{1.0}
{\small
\begin{split}
&\text{ \ }\Delta w_{i}=\begin{cases}
-a^{-}w_{i}(1-w_{i}),\ \ \text{if}\ \ t_{out}-t_{in}<0\\
+a^{+}w_{i}(1-w_{i}),\ \ \text{if}\ \ t_{out}-t_{in}\geq0 
\end{cases}\\
&w_{i}\leftarrow w_{i}+\Delta w_{i}
\label{stdp}%
\end{split}
}
\end{equation}
$t_{in}$ and $t_{out}$ are the spike times of the pre-synaptic (input) and
the post-synaptic (output) neuron, respectively. If the $i^{th}$
input neuron spikes before the output neuron spikes, the weight $w_{i}$ is
increased; otherwise the weight is decreased.\footnote{The input neuron is
assumed to have spiked \emph{after} the output neuron spiked.} Learning refers
to the change $\Delta w_{i}$ in the (synaptic) weights with $a^{+}$ and
$a^{-}$ denoting the learning rate constants. These rate constants are
initialized with low values $(0.004,0.003)$ and are typically increased for 
every 1500 input images as learning progresses \cite{Kheradpisheh_2016}. This STDP rule is considered
simplified because the amount of weight change doesn't depend on the time 
duration between pre-synaptic and post-synaptic spikes. In this work, backpropagation is used only
in the classification layers (L3-L4-L5) of the network with a single hidden
layer L4.
\vspace{-0.4cm}
\section{Continuous Learning}
By continuous learning we mean that the network in Figure \ref{network1} will start with two output neurons in L5 and be trained to classify the digits $\{0,1\}$. During this training the error is backpropagated from layer L5 only as far as L3. After this training is complete two new neurons will be appended to the L5 layer and then trained to classify the digits $\{2,3\}$. This is continued in the same manner for the three remaining classes $\{\{4,5\},\{6,7\},\{8,9\}\}$. We proceed in the rest of this section to give the details of this training by specifying the cost function along with the (cost per synapse) weight regularizer. The neural network in this work has a softmax output layer which is the likelihood of the input image belonging to a particular class. Let $\textbf{X} \in R^{3630}$ denote the (flattened) spike features in L3 and ${\theta} \in R^{1500 \times 3630}$ denote the weights from L3 to the hidden layer L4. For task 1 there are two output neurons and we let $C_{1}$ denote the cross-entropy cost computed with the softmax outputs of these two neurons. For task 2 there are now four output neurons and we let $C_{2}$ denote the cross-entropy cost computed with the softmax outputs of these four neurons. The costs $C_{3}, C_{4}, C_{5}$ are defined in a similar manner. The L4 and the L5 weights are updated using SGD on mini-batches. $C^{(m)}_{1}$ denotes the cost of the $m^{th}$ input mini-batch for the task $\{0,1\}$. $C^{(m)}_{2},...,C^{(m)}_{5}$ are similarly defined. During training for task 1 the weights ${\theta} \in R^{1500 \times 3630}$ are updated as usual according to
\vspace{-0.15cm}
\begin{equation}
\scalefont{1.0}
\Delta\theta = - \eta\frac{\partial C_1^{(m)}}{\partial \theta}. \label{deltatheta}
\end{equation}
After training is completed for task 1, we need to know the  importance of each of the weights ${\theta}_{rs}$ for $r=1,...,1500$ and $s=1,....,3630$ in terms of classifying the images of task 1. This is necessary because when we proceed to train on task 2 these "important" weights should not be allowed to change significantly. That is the network must be forced to use the other weights for the training of task 2. Accordingly, we next define a \emph{cost per synapse} regularizer during the training of task 2 to help prevent changes to the so called important weights of the task 1. The change in the cost per each synapse $\Delta C_{1,{rs}}^{(m)}$ is defined as
\vspace{-0.2cm} 
\begin{equation}
\scalefont{1.0}
\Delta C_{1,{rs}}^{(m)} \triangleq \frac{\partial C_{1}^{(m)}}{\partial\theta_{rs}}%
\Delta\theta_{rs}=-\eta\left(  \frac{\partial C_{1}^{(m)}}{\partial\theta
_{rs}}\right)^{2} \label{deltac}
\end{equation}
with
\begin{equation}
\scalefont{1.0}
\Delta C_{1}^{(m)} \triangleq\left\{  \Delta C_{1,rs}^{(m)}\right\}
_{\substack{r=1,...,1500\\s=1,...,3630}}\in%
\mathbb{R}
^{1500\times3630}%
\end{equation}
For each task there are $M$ mini-batches with $P$ images per mini-batch for a total of $N=MP$ input images for each task. The average change in cost for $\theta_{rs}$ is given by
\vspace{-0.2cm}
\begin{equation}
\scalefont{1.0}
\small{
f_{1,{rs}} \triangleq  \frac{1}{M}{\sum_{m=1}^{M}\Delta C_{1,{rs}}^{(m)}} = - \eta\frac{1}{M}{\sum_{m=1}^{M}{\left(\frac{\partial C_1^{(m)}}{\partial \theta_{rs}}\right)}^2} \label{sumdeltac}
}
\end{equation}
with
\begin{equation}
\scalefont{1.0}
 f_{1} \triangleq\left\{  f_{1,rs}\right\}
_{\substack{r=1,...,1500\\s=1,...,3630}}\in%
\mathbb{R}
^{1500\times3630} \label{substackf}
\end{equation}
A softmax output layer with a cross-entropy cost function and one-hot encoded labels is the same as the log-likelihood cost function \cite{Nielsen}. The MNIST label $l$ with $l \in \{0,1,2,...,9\}$ corresponds to the $k^{th} (=l+1)$ output neuron with $k \in \{1,2,...,10\}$, respectively. Let $\mathbf{X}^{(m)}=\{(X^{(im)},l_{i}),i=1,...,P\}$ denote the images and corresponding labels in the $m^{th}$ mini-batch. In Equation (\ref{sumdeltac}) the average cost $C_{1}^{(m)}$ for mini-batch $m$ is 
\vspace{-0.2cm}

\begin{equation}
\scalefont{1.0}
{\small
\begin{split}
C_{1}^{(m)} = \frac{1}{P}\sum_{i=1}^{P}C_{1}(\textbf{X}^{(im)})= -\frac{1}{P}\sum_{i=1}^{P}\sum_{k=1}^{n_o=2}y_{k}\ln a_{k}^{L5}(\textbf{X}^{(im)})\\= -\frac{1}{P}\sum_{(\textbf{X}^{(im)},l_i)\in \textbf{X}^{m}}\ln a_{l_i+1}^{L5}(\textbf{X}^{(im)}) \label{loglik}
\end{split}
}
\end{equation}
as $y_{l+1}=1$ and $y_{k}=0$ for $k \neq l+1$. Here L5 indicates the last layer and $a^{L5}(\textbf{X})$ indicates softmax output activations. Substituting Equation (\ref{loglik}) in to Equation (\ref{sumdeltac}), Equation (\ref{substackf}) becomes
\vspace{-0.2cm}

\begin{equation}
\scalefont{1.0}
f_1  = - \eta\frac{1}{M}{\sum_{m=1}^{M}{\left( {\frac{\partial C_{1}^{(m)}}{\partial \theta}}\right)}^2} \in%
\mathbb{R}
^{1500\times3630} \label{sumdeltac2}
\end{equation} 
In \cite{Kirkpatrick2017} the authors state that near a minimum of the cost the $(r,s)$ component of Equation (\ref{sumdeltac2}) given by $-\dfrac{1}{M}\sum\limits_{m=1}^{M}\left(  \dfrac{\partial C_{1}^{(m)}%
}{\partial\theta_{rs}}\right)  ^{2}$%
is the same as 
\begin{equation}
I_{rs}(\theta)\triangleq\dfrac{1}{M}\sum\limits_{m=1}%
^{M}\dfrac{\partial^{2}C_{1}^{(m)}}{\partial\theta_{rs}^{2}},
\end{equation}
with some limitations \cite{kunstner2019} and is the
Fisher information \cite{chiasson_2013} for the parameter $\theta_{rs}.$ $I_{rs}(\theta)$ is a
measure of the \textquotedblleft importance\textquotedblright\ of the weight
$\theta_{rs}.$ A large value of $I_{rs}(\theta)$ implies that small changes in
the value of $\theta_{rs}$ will lead to a large increase in average cost
(classification error). When the network is to train for task 2, those weights $\theta_{rs}$ with a large $I_{rs}(\theta)$ computed from task 1 must now be constrained to only small changes so the network will continue to classify the images of task 1 correctly. That is, when
training on task 2, the network must be forced to (essentially) use only those weights that
had a small value of $I_{rs}(\theta)$ from task 1.
So, the cost per synapse for the first task $f_1$ (calculated during the last epoch of training for the first task) gives the relative importance of the weights for the task1 classification problem. 

Let $\Delta{\theta_1} \in%
\mathbb{R}
^{1500\times3630}$ be the change in weights during the last epoch of task 1. Further $\hat{\theta}_{1}$ denotes the value of the weights after training on task 1. The second task is trained using the regularized cost function given by 
\begin{equation}
\scalefont{1.0}
C_{2}^{reg} \triangleq C_{2} + \frac{\lambda}{2N}({\theta}_{2}-{\hat{\theta}}_{1})\odot F_{1}\odot({\theta}_{2}-{\hat{\theta}}_{1})
\end{equation}
with
\begin{equation}
\scalefont{1.0}
F_{1} \triangleq f_1 \oslash (\Delta{\theta_1}\odot\Delta{\theta_1} + \xi)  
\end{equation}
where $\odot$ and $\oslash$ represent the Hadamard product and division, respectively. $\xi$ is a small positive number added to each element of the matrix to prevent division by zero when doing Hadamard division. Similarly, $f_{t}$ is calculated during the last epoch of training task $t$, ${\Delta\theta}_{t}$ denotes the change in the weights during the last epoch, and finally $\hat{\theta}_{t}$ denotes the weights at the end of training task $t$. Task $t$ is trained by adding a weight regularizing term to prevent the "important" weights from the previous tasks being changed significantly. With $f_{0} \in%
\mathbb{R}^{1500\times3630}$ a matrix of zeros define 
\vspace{-0.2cm}
\begin{equation}
\scalefont{1.0}
F_t \triangleq \sum_{\tau=0}^{t-1}f_\tau \oslash (\Delta{\theta_\tau}\odot\Delta{\theta_\tau} + \xi). 
\end{equation}
then we can write the cost function of the $t^{th}$ as
\vspace{-0.2cm}
\begin{equation}
\scalefont{1.0}
    C_{t}^{reg} = C_{t} + \frac{\lambda}{2N}\odot F_{t} \odot({\theta}_{t}-{\hat{\theta}}_{t-1})\odot({\theta}_{t}-{\hat{\theta}}_{t-1}) , \hspace{4mm}  \hspace{2mm} \text{t=1,2,3,4,5} \label{newcost}
\end{equation}
$\hat{\theta}_{t-1}$ are the weights between L3 and L4 layers at the end of $(t-1)^{th}$ task and $N=MP$ is the number of input images. \textbf{Remark} Note that only the weights connecting L3 to L4 are subject to cost per synapse regularization. The weights connecting L4 to L5 are trained without regularization and use the AR1 method to train sequentially  \cite{MALTONI2019}. The parameter updates for the cost function are
\begin{equation}
\scalefont{1.0}
\small{
\frac{\partial C_t^{reg}}{\partial {\theta_t}}=\frac{\partial C_t}{\partial {\theta_t}}+
\frac{\lambda}{N} F_t \odot ({\theta_t}-{\hat{\theta}}_{t-1}) \label{newdcdt}
}
\end{equation}

In \cite{zenke2017} the cost per synapse is calculated over all the
\begin{wrapfigure}[7]{l}{0.5\linewidth}
\vspace{-0.55cm}
\centering
\includegraphics[width=1.1\linewidth, height=0.65\linewidth]{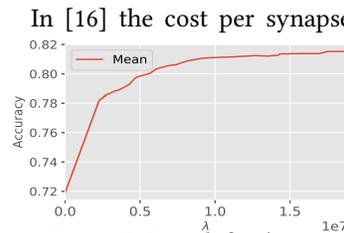}
\vspace*{-1.0cm}
\captionof{figure}{Search for $\lambda$.}
\label{lambda_vs_acc}
\end{wrapfigure}
  training epochs (rather than just the last epoch). The parameter $\lambda$ in Equation (\ref{newdcdt}) was optimized with validation data. Figure \ref{lambda_vs_acc} shows the effect of $\lambda$ on accuracy. Results for each $\lambda$ were obtained from $10$ different  weight initializations. 
\vspace{-0.4cm}
\subsection{Results with Disjoint Tasks}
In this section the network was not presented with any
\begin{wrapfigure}[7]{l}{0.5\linewidth}
\vspace{-0.6cm}
\centering
\includegraphics[width=1.1\linewidth, height=0.7\linewidth]{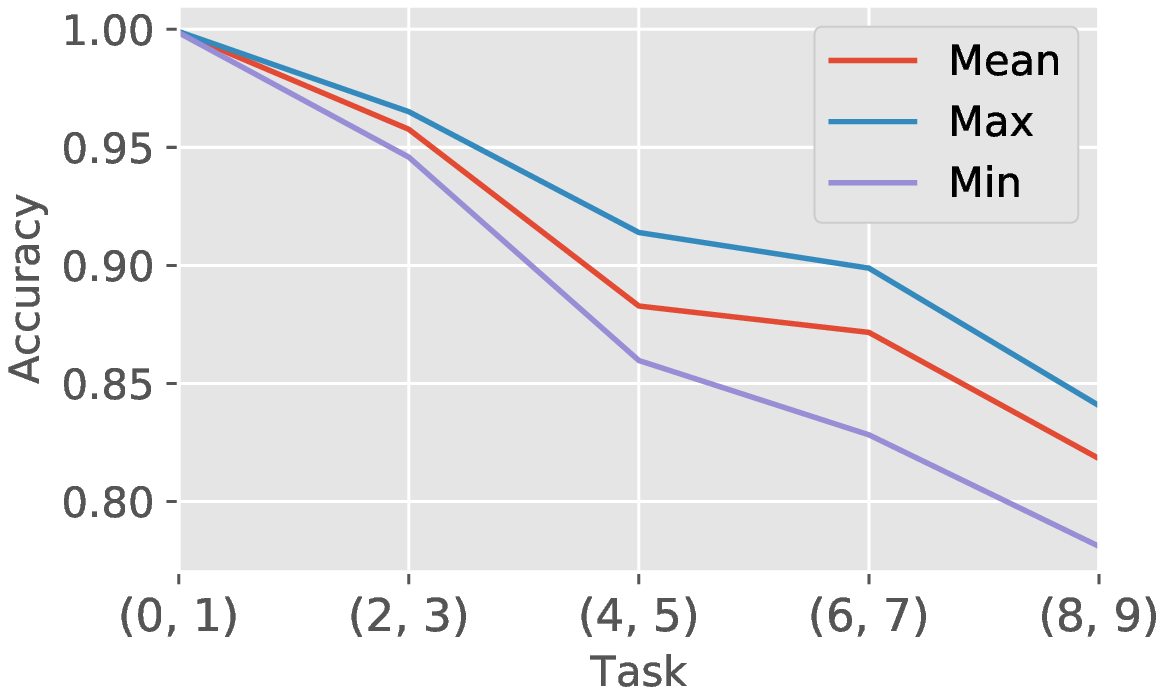}
\vspace*{-1.0cm}
\captionof{figure}{Test accuracy}
\label{disjoint_accuracy}
\end{wrapfigure}
 of the data from the previous tasks. Figure \ref{disjoint_accuracy} shows the trend of testing accuracy as the network is trained on disjoint tasks with 10 different weight initializations. The highest testing accuracy achieved for this disjointly trained tasks was $84.61\%$. $\lambda$ was set to $2.03\times10^7$. in Figure \ref{disjoint_accuracy} 'Max' in the legend indicates the weight initialization that resulted in highest test accuracy and 'Min' indicates the weight initialization that resulted in lowest test accuracy.  
\vspace{-0.4cm}
\subsection{Results by Replaying}
Here the data for the new task was expanded with $10\%$ of 
\begin{wrapfigure}[8]{l}{0.5\linewidth}
\vspace{-0.6cm}
\centering
\includegraphics[width=1.1\linewidth, height=0.7\linewidth]{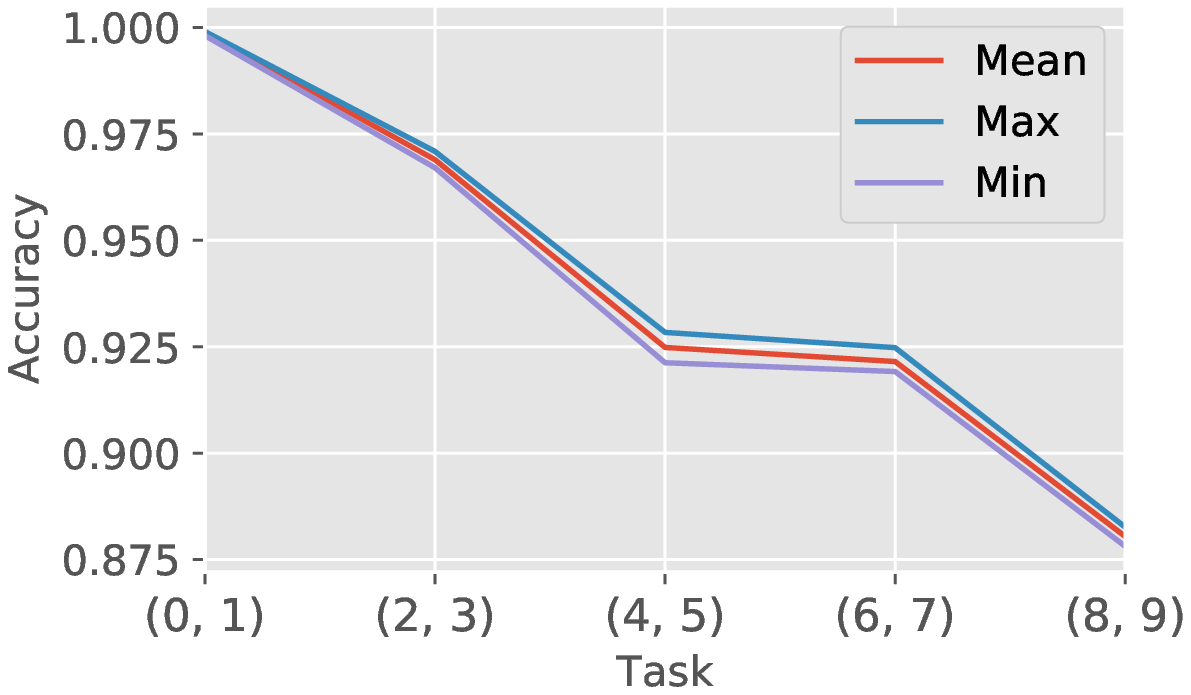}
\vspace{-1.0cm}
\captionof{figure}{Test accuracy.}
\label{buffered_accuracy}
\end{wrapfigure}
 data from all the previously trained tasks. The accuracy trend as the learning progresses is shown in the Figure \ref{buffered_accuracy} for 10 different weight initializations with $\lambda$ same as the disjoint case. The highest accuracy achieved by replaying $10\%$ of the data from the previous tasks was $88.41\%$. For all of the above reported experiments the hyper-parameter $\eta = 1.0 \times 10^{-3}$, the mini-batch size $P = 10$ and the number of mini-batches per task $M = 1000$.

\section{Conclusion}
We achieved a final accuracy of $88.4\%$ in sequential training of MNIST dataset. While we do not claim state-of-the-art classification accuracy we proposed a hybrid method that combines bio-inspired feature extraction and the cost per synapse metric that is scalable to more challenging datasets. This is a work in progress and we are currently working towards extending this work to a more challenging EMNIST dataset \cite{emnist} by using surrogate gradients \cite{vaila2020} that can potentially eliminate floating point matrix-vector multiplications when implemented in a custom hardware.  
\vspace{-0.4cm}
\bibliographystyle{ACM-Reference-Format}
\bibliography{icons_2020}

%

\end{document}